\documentclass[review]{elsarticle}

\usepackage[normalem]{ulem}
\usepackage{multirow}
\usepackage{mathtools}

\usepackage{amsthm,amssymb,amsmath,layout}
\usepackage{algorithm}

\usepackage[noend]{algpseudocode}
\usepackage{setspace}

\usepackage{tikz}
\usetikzlibrary{arrows,calc,patterns,positioning,shapes}
\usetikzlibrary{decorations.pathmorphing}

\tikzset{
modal/.style={>=stealth',shorten >=1pt,shorten <=1pt,auto,node distance=1.5cm,
semithick},
unworld/.style={circle,draw,minimum size=0.5cm,fill=red!15},
world/.style={circle,draw,minimum size=0.5cm,fill=gray!15},
point/.style={circle,draw,inner sep=0.5mm,fill=black},
reflexive above/.style={->,loop,looseness=7,in=120,out=60},
reflexive below/.style={->,loop,looseness=7,in=240,out=300},
reflexive left/.style={->,loop,looseness=7,in=150,out=210},
reflexive right/.style={->,loop,looseness=7,in=30,out=330},
coil/.style={decorate, decoration={coil,amplitude=4pt,segment length=5pt}},
snake/.style={decorate, decoration={snake}},
zigzag/.style={decorate, decoration={zigzag}}
}

\usepackage{listings}

\usepackage{color,soul}

\theoremstyle{plain}

\newtheorem{thm}{Theorem}[section]

\theoremstyle{definition}

\newtheorem{rem1}{Remarks}[section]
\newtheorem{rem}[rem1]{Remarks}
\newtheorem{exa}[thm]{Example}

\newtheorem{defi1}{Definition}[section]
\newtheorem{defi}[defi1]{Definition}

\newtheoremstyle{defC}%
  {6pt}
  {6pt}
  {\normalfont}
  {}
  {\bfseries}
  {{\bfseries .}}
  {5pt plus 1pt minus 1pt}
  {\thmname{#1} \thmnumber{#2} \thmnote{\normalfont#3}}
\theoremstyle{defC}

\newtheoremstyle{thmC}%
  {6pt}
  {6pt}
  {\itshape}
  {}
  {\bfseries}
  {{\bfseries .}}
  {5pt plus 1pt minus 1pt}
  {\thmname{#1} \thmnumber{#2} \thmnote{\normalfont#3}}

\theoremstyle{thmC}

\numberwithin{equation}{section}

\journal{arxiv}

\begin{document}

\begin{frontmatter}

\title{Meet MASKS:\\ A novel Multi-Classifier's verification approach}

\author{Amirhoshang~Hoseinpour~Dehkordi}
\address{School of Computer Science, Institute for Research in Fundamental Sciences, Tehran, 19395-5746 Iran}

\author{Majid~Alizadeh\corref{mycorrespondingauthor}}
\address{School of Mathematics, Statistics and Computer Science, College of Science, University of Tehran, Tehran, 14155-6455 Iran}

\cortext[mycorrespondingauthor]{Corresponding author}
\ead{majidalizadeh@ut.ac.ir}


\author{Ali~Movaghar}
\address{Department of Computer Engineering, Sharif University of Technology, Tehran, 11155-9517 Iran}





\begin{abstract}
In this study, a new ensemble approach for classifiers is introduced. A verification method for better error elimination is developed through the integration of multiple classifiers.
A multi-agent system comprised of multiple classifiers is designed to verify the satisfaction of the safety property.  
In order to examine the reasoning concerning the aggregation of the distributed knowledge, a logical model has been proposed. 
To verify predefined properties, a Multi-Agent Systems' Knowledge-Sharing algorithm (MASKS) has been formulated and developed. 
As a rigorous evaluation, we applied this model to the Fashion-MNIST, MNIST, and Fruit-360 datasets, where it reduced the error rate to approximately one-tenth of the individual classifiers.
\end{abstract}


\end{frontmatter}


\section*{Introduction and background}\label{S:one}
Nowadays, classifiers are a mandatory part of most real-world AI applications. 
Among these, and in particular, are safety-critical systems, in which failures may result in fatal consequences. 
The most common types of classifiers are based on statistical methods employed to improve the performance of certain tasks (i.e., Artificial Neural Networks (ANNs)). 
Such an AI-assisted statistical process of decision-making is based on the acquired information. This complicates the interpretation of the cause underlying the decisions made.
As pertaining to performance measures, two issues are material and often cause vulnerabilities. First, are architectural flaws and deficiencies, and second, having limited sets for training the classifiers. Besides architectural flaws and limited datasets, the system may even be confronted with prearranged noisy inputs, which have been inputted with malignant intentions (i.e., adversarial examples \cite{biggio2013evasion} and \cite{szegedy2013intriguing}).  For example, in image processing cases, adversarial inputs tend to locate themselves towards neighbourhoods (which, considering predefined norms, are images located in the near vicinity). Furthermore, towards manipulations (which, in human perception, are considered visually similar), such as camera angle changes and image skewing \cite{Huang0}. In this case, it seems the property proposed to describe the classifier's knowledge should incorporate possible neighbourhoods and manipulations in order to verify its correct performance. Here, and considering formal verification, theoretical understanding regarding the correctness of our formula (property) gains importance. 
So that, verification methods have been applied more widely to classifiers \cite{tornblom2020formal}. As a basis, one can reference the safety verification on ANNs (as classifiers) in \textit{Multi-Layer Perceptrons Neural Networks}, in which linear computation constraints are abstracted to ``Boolean combinations of linear arithmetic constraints'', Pulina \textit{et al.} \cite{Pulina0}.
Advancing in time, environmental modelling, formal specification, system modelling, computational engines, and correct-by-construction design were further considered, Sanjit \textit{et al.} \cite{Seshia0}. 
Building on the aforementioned studies, a unified framework was created in order to provide safe and reliable integration for human-robot systems, D. Sadigh  \textit{et al.} \cite{Sadigh0}.
In addition, and regarding safety-critical verification of ANNs (as classifiers), Scheibler \textit{et al.} applied a bounded model checking method. In this work, the formulas introduced by the verification method were solved via iSAT3 (an SMT-solver) and special deductions \cite{Scheibler0}.
Following their research, improvements were made by Katz \textit{et al.} by taking into consideration more general properties. This allowed them to verify networks by using simplex methods which include piece-wise linear ReLU activation functions \cite{Katz0}. An approach to studying the robustness of multiple classifiers with a focus on ANNs is developed by Gross \textit{et al.} ``to generate optimal results for medium-sized neural networks'' \cite{gross2020robustness}.
Albeit, none of these methods could verify larger networks, such as Alexnet, which includes about $\sim$650,000 ReLU nodes\cite{krizhevsky2012imagenet}. All of these methods put effort into the verification of the model, so the time complexity was based on the size of the model. Therefore, and due to the growing rate of size of such models, these models could not be applied to verify the latest models.

Pushing onward, and as another perspective, point-wise robustness (which could be applied for each layer of ANNs -as classifiers) was introduced and inserted into the verification method, Huang \textit{et al.} \cite{Huang0}.  Their algorithm succeeded in exhaustively searching the neighbourhood of a network's inputs with reasonable complexity and in an acceptable time frame. This model appears well optimised and more general. In their research, the results were impressive by adequately defining neighbourhoods and manipulating them as properties (they could get rid of misclassifications). However, defining good neighbourhoods and manipulation in general for classifiers has difficulties. Besides, it is not designed to verify safety-critical cases regarding a collaborative system of classifiers because it cannot reason about the knowledge generated with multiple classifiers.

In this study, a new ensemble approach for classifiers is introduced.
Unlike statistical ensemble approaches, this method can determine the source of knowledge in a deterministic manner. Moreover, we will examine: ``Could this multi-agent scenario helps us to reduce error in the case of improper neighbourhoods and manipulations?''. 
Next, according to the predefined properties, a set of all possible “manipulations” is collected. Consequently, all inputs in the test set plus all manipulations would be fed into all classifiers as inputs. Each set of classifiers would output a set of classes.

The structure of the paper is as follows. In section \ref{sec:Epistemic}, the developed model, based on Epistemic logic and a multi-agent system is presented. In section \ref{sec:Interactions}, collaboration algorithms are introduced. In section \ref{sec:EarlyResults}, two
types of examples explaining the details of the proposed model are provided.  Finally, section \ref{sec:Conclusion} concludes the paper.

\section{ Logical Model for classifiers}
\label{sec:Epistemic}
In this section, we introduce a novel interpretation of dynamic epistemic logic that suits the formal description of the dynamics of the knowledge of classifiers as the agents in a Multi-Agent System.
\subsection{Safety in Image Verification}
Formal verification is the mathematical process to ensure that a model fulfills some properties in an environment \cite{bjesse2005formal}. Pulina \textit{et al.}, \cite{Pulina0} defined a verification method for classifiers using \textit{safety} as the property to be satisfied by the classifiers. \textit{Safety} is defined as a consistent classification result due to minor input changes.
\par
Let us illustrate the property of Safety by a sample example. Classification is a process in which an input space is partitioned into a number of output classes. Assume we have an image classifier that accepts  $100 \times 100 $ matrices representing grayscale input images, and the classifier is going to verify if an image has a face or not. So the input space is a $10000$-dimensional vector space, and the output space would consist of two classes: ``images with face'' and ``images without face''.
Note that by changing the value of a single pixel of an image, the input image \textit{changes} but human perception of an image does not alter by such minor changes. On the other hand, consider the objects that are recognised as being the same through various camera angles, or at night or day, etc.; the human mind perceives them as the same object, whereas they may be completely different as an input point for the classifier. To verify safety, these minor changes shall not lead to inconsistent classification results.\\
To overcome these difficulties, Huang \textit{et al.}, \cite{Huang0}, introduced the concepts of \textit{Neighbourhood} and \textit{Manipulation} sets of an input image. 
\begin{defi}
Let $X\subseteq\mathbb{R}^n$ be an input domain and $x_0\in X$, then the {\em $\epsilon$-neighbourhood} of $x_0$ is the set: 
$$\eta_\epsilon(x_0)=\{ x \in X \mid d(x_0,x)<\epsilon \},$$
  where $d$ is an arbitrary distance metric.
\end{defi}
\begin{defi} Let $X\subseteq\mathbb{R}^n$ be an input domain and $x_0\in X$, then the {\em $\mathcal{F}$-manipulation} of $x_0$ is the set: 
$$\mu_\mathcal{F}(x_0)=\{ x \in X \mid \mathcal{F}(x_0,x)=1 \},$$
  where $\mathcal{F}:\mathbb{R}^n\times\mathbb{R}^n \rightarrow \{0,1\}$ as an arbitrary function.
  \end{defi}
  Whenever $\epsilon$ and $\mathcal{F}$ is clear from the context, we will drop the subscripts in $\eta $ and $\mu $ and will call them, neighbourhood and manipulation set.
\begin{rem}
The verification methods goals are developed to satisfy specific predefined properties. However, the classification generally aims to reduce errors. The verification methods could help classifiers to be self-aware, if properties are defined correctly. Nevertheless, defining such properties is difficult in many cases.
As we can see, Huang \textit{et al.}, have presupposed an arbitrary but fixed measure of distance $\epsilon$ for neighbourhood and characteristic function of manipulation set $\mathcal{F}$, but defining the $\epsilon$ for neighbourhood and the function $\mathcal{F}$ in manipulation is not trivial. Examples \ref{exam2}, \ref{exam3}, and \ref{exam4} will show how multi-classifiers could reduce the misclassification caused by improper safety properties.
\end{rem}
Let $G$ be a image classifier. For an input $x$, $[x]_G$ is the label of output class of $x$ in $G$. For a set of input points $X$, $[X]_G$, which is defined as $\bigcup_{y\in X}[y]_G$, is the label of output class of $X$ in $G$.

The concepts of safety and robustness are defined in Huang \textit{et al.} \cite{Huang0}. We explain these within the literal context of this paper in the definitions below.
\begin{defi}
Let $G$ be a classifier. We say that $G$ is safe for input point $x_0$ exactly when for all input point $x$,  
$ x \in \eta(x_0)\cup\mu(x_0)$ implies that $ [x]_G=[x_0]_G$.
\end{defi}
\begin{defi}Let $G$ be a classifier and $x$ be an input point.  $x$ is called {\em robust} exactly when  $[\eta(x) \cup \mu(x)]_G=[x]_G$.
\end{defi}
\begin{defi}\label{def:verified}A property  $\rho(x)$ in a given  classifier $G$ is {\em verified} exactly when  $|[\rho(x)]_G|=1$.\\
In which ``$|A|$'' is size of the set $A$. 
\end{defi}
For a classifier $G$ and input point $x_0$, if every point in the $\eta(x_0) \cup \mu(x_0)$ is classified into the same class, the input point $x_0$ would be a robust point and the safety property would be verified for $x_0$.
\begin{defi}\label{def:verifiedMAS} A property  $\rho(x)$ in a given set of classifiers $A_G = \{ G_1, \dots , G_n \}$ is {\em verified} exactly when  $|\bigcap_{G \in A_G} [\rho(x)]_G|=1$.
\end{defi}
\subsection{Dynamic Epistemic Logic and Public Announcement Logic}
\newcommand{\K}[1][]{\mathsf{K}_#1}
\newcommand{\D}[1][]{\mathsf{D}_#1}
\newcommand{\PA}[1]{\mathsf{[#1]}}
Epistemic Logic is a modal logic that is concerned about knowledge and reasoning about knowledge. The basic modal operator of epistemic logic, written as $\K{}$, is read as "It is known that,".
Dynamic Epistemic Logic is introduced to model the epistemic interactions of the agents, when dealing with more than one agent.\\
\newcommand{\DEL}{\mathsf{DEL}}
\newcommand{\PAL}{\mathsf{PAL}}
Here we review the basics of Dynamic Epistemic Logic ($\DEL$) to be able to reason regarding agents' knowledge. $\DEL$ is a logical framework designed to deal with the dynamics of knowledge of agents in multi-agent systems by adding dynamic modalities to epistemic logic.
Hence, the agent's actions can alter the facts of the possible worlds. For more details, see \cite{van2006logics, balbianibaltagditmarschherzighoshidelima2008, sep-logic-epistemic, sep-dynamic-epistemic, sep-possible-worlds}. The most simple version of Dynamic Epistemic Logic is Public Announcement Logic ($\PAL$), which is an extension of multi-agent epistemic logic, where dynamic operators are employed to model the epistemic consequences of announcements to the entire group of agents (see \cite{plaza1989logics, plaza2007logics, van2007comments}). 
\begin{defi}[Language of $\PAL$]
Let $\Phi$ be a set of propositional variables and $Ag=\{1,\ldots,n\}$ be a finite set of agents.
The epistemic language is defined inductively through the following grammar, in BNF:
$$\phi \Coloneqq \top \mid p \mid \neg\phi \mid (\phi\land\phi) \mid 
\K{i}\phi \mid \D{A} \phi \mid \PA{\phi}\phi. $$
where $p\in \Phi$, $i \in {Ag}$ and $A\subseteq {Ag}$. 
\end{defi}
The intended meaning of $\K{i}\phi$ is  ``agent $i$ knows $\phi$''. $\D{A}\phi$ is the distributed knowledge of $\phi$ among agents in $A$, i.e., if the agents pulled their knowledge altogether, they would know that $\phi$ holds. The formula $\PA{\psi}\phi$ means that after a truthful announcement of $\psi$, $\phi$ holds. \\
An inference system for PAL can be found in \cite{wang2013axiomatizations}.\\
For semantics, we will use Kripke models. A Kripke model for epistemic logic is tuples
$\mathcal{M}=(W,R_1,\cdots,R_n,V)$, where $W=\{w_0, w_1,\cdots,w_{k}\}$ is a set of worlds or states, $R_i\subseteq W\times W$ is the \textit{equivalence} relation for every agent $i$, and $V: W \rightarrow 2^\Phi$ is the evaluation function. The fact that $R_i$ is an equivalence relation means that the agent $i$ cannot tell states $w$ and $w'$ apart, when $w R_i w'$.\\
As usual, the truth of complex formulas are defined inductively:
\begin{defi}
Let $\mathcal{M} = (W, R_1, \dotsc, R_n, V)$ be model for $\PAL$. We say that $\phi$ is true in $w\in W$, written $\mathcal{M}, w\models \phi$, when:
\begin{itemize}
    \item 
$\mathcal{M},w\models p$\quad iff \quad $p\in V(w)$;
\item
$\mathcal{M},w\models \neg\phi$ \quad iff\quad
 $\mathcal{M},w\nvDash\phi$;
 \item
$\mathcal{M},w\models \phi\land\psi$\quad iff \quad 
$\mathcal{M},w\models\phi\textrm{~and~}\mathcal{M},w\models\psi$;
\item
$\mathcal{M},w\models \K{i} \phi$\quad iff \quad
$\forall v\in R_i(w),\  \mathcal{M},v\models\phi$, where $R_i(w) = \{ w'\mid w R_i w'\}$;
\item
$\mathcal{M},w\models \D{A}\phi$ \quad iff
\quad $\forall v\in R_{\D{A}} (w),\  \mathcal{M},v\models\phi$, where $R_{\D{A}}:=\underset{i\in A}{\bigcap} R_i$;  
\item
$\mathcal{M},w\models\PA{\psi}\phi$\quad iff\quad
$\mathcal{M},w\models\psi$ implies  $\mathcal{M}^{\psi},w\models\phi$,
\end{itemize}
where $\mathcal{M}^{\psi}$ is the updated Kripke model $\mathcal{M}$ by the announcement $\psi$ in which $W^{\mathcal{M}^\psi}:=\{ w\in W\mid (\mathcal{M}, w)\models \psi\}$, the relation $R_i^{\mathcal{M}^\psi}:= R_i\cap(W^{\mathcal{M}^\psi}\times W^{\mathcal{M}^\psi}) $ and  the valuation $V^{\mathcal{M}^\psi}$ is the restriction of the valuation $V$ to $W^{\mathcal{M}^\psi}$\cite{wang2011public}. 
\end{defi}
Here, we concentrate on the interplay between knowledge and epistemic action, which only modifies the agents' knowledge while leaving the facts unchanged. So, $\PA{\psi}$ is an operator that takes us to a new model consisting only of those worlds where $\psi$ has been rendered as true. Therefore, after the announcement of $\psi$, no agent considers worlds where  $\psi$ was false. Hence, we should evaluate formulas in the sub-model $\mathcal{M}^{\psi}$.\\
Note that the operator $\D{A}$ can be interpreted as a necessity operator of the relation on $R_{\D{A}}$.\\ 
If we consider all such Kripke models, the set of all valid formulas obtained from these semantics is known as modal logic ${\bf S5}$, see \cite{Ditmarsch0}.

\subsection{Classifiers and Dynamic Epistemic Logics }
\label{sec:synsem}
Let us consider a classifier $G$, for which we are going to verify safety. Here we use safety (as the property) to simplify the demonstration, but any property could be defined arbitrarily. Let $x$ be an input point. Suppose that the $\epsilon$ and $\mathcal{F}$ for defining neighbourhood and manipulation sets are given. As defined in definition \ref{def:verified}, safety of $x$ is verified in $G$, exactly when $x$ is robust for $G$. 
Suppose that input point $x$ is not robust, and some points in its neighbourhood and manipulation set are classified into different classes. These classes can be considered as alternative outputs. One can reduce these cases by employing multiple classifiers within a knowledge sharing multi-agent system.\\
Suppose we have a group $\mathbb{G}$ of classifiers which are sharing knowledge about an input point. In case the intersection of these output classes (possible knowledge) is a single output class, such a system can verify properties by applying the shared knowledge. Note that safety of $x$ is verified in $\mathbb{G}$,  exactly when $\bigcap_{G \in \mathbb{G}} [x]_G$ is a singleton, cf. definition \ref{def:verifiedMAS}.
\par To formalise sharing of knowledge by classifiers, we introduce The $\DEL$- model
\newcommand{\mm}{\mathfrak{M}_\mathbb{G}}
 $\mm = (W, R_1, \dotsc, R_n, V) $ of $\PAL$ as follows.

Suppose $\mathbb{G} = \{G_1, \dotsc, G_n\}$ is a finite set of classifiers, $X$ is the set of input points, and $C$ is the set of all output classes. The set of propositional variables $\Phi$ will be interpreted by $X \times C$, i.e., each propositional variable is interpreted by a a pair $(x,c)$, where $x$ is an input point and $c$ is a  output class.
Each classifier will represent one agent in $\PAL$.\\
The set of worlds $W$ is all possible output results for the input values plus an extra world $\Bar{c}$ i.e.,  
$$W \coloneqq \{c \in C \mid \exists x\in X\ \exists G \in \mathbb{G}\ \text{such that}\ c \in [\eta(x)\cup\mu(x)]_G\}\cup\{\Bar{c}\}.$$

We add  $\Bar{c}$ to the set of worlds in order to find out ``which class appears as an output of any agent?''.
We also notice that, since classification of distinct input points are not related to each other, we can consider a fixed input point $x$ and create the model based on that. The final model will be the disjoint union of the models for each input.\\
So, suppose that $x$ is a given fixed input point, then for any given $c\in C$ we define $R^x_i(c)$  and $V^x(c)$ abbreviate $R_i(c)$  and $V(c)$, respectively as follows:
\begin{equation*}
R_i(c) \coloneqq
\begin{cases}
\{c\} & c \notin [\eta(x)\cup\mu(x)]_{G_i} \\
\{c' \mid c' \in [\eta(x)\cup\mu(x)]_{G_i}\}\cup\{\Bar{c}\}\ &  c \in [\eta(x)\cup\mu(x)]_{G_i}
\end{cases}
\end{equation*}
$$V(c) \coloneqq \{(x,c) \mid \exists G \in \mathbb{G}\ \text{such that}\ c\in [\eta(x)\cup\mu(x)]_G\}.$$

$R_i(\Bar{c})$  and $V(\Bar{c})$ are defined as follows:
$$R_i(\Bar{c}) \coloneqq \{c' \mid c' \in [\eta(x)\cup\mu(x)]_{G_i}\}{\cup\{\Bar{c}\}}, $$ 
$$V(\Bar{c}) \coloneqq \{(x,c) \mid \exists G \in \mathbb{G}\ \text{such that}\ c\in [\eta(x)\cup\mu(x)]_G\}.$$


Note that since the above definitions are given for a single  fixed input $x$, then for any given $c\in C$, we have $V(c)=\{(x,c)\}$ or $V(c) = \varnothing $. 

Now we can show that a point is robust for a classifier if and only if its interpretation is valid in $\mm$.
\begin{thm}
Suppose that $\mathbb{G} = \{G\}$ is a single-classifier system, $x$ is an input point and $c$ is an output class of $G$. Then
\begin{center}
    $[\eta(x)\cup\mu(x)]_{G} = \{c\}$ if and only if $\mm, c \models \K{}p_c$,  where $p_c$ is interpreted by $(x,c)$.
\end{center}
\end{thm}
\begin{proof}
Only-if part.
Suppose $\mm, c \models \K{}p_c$, where $p_c$ is interpreted by $(x,c)$. As we have only one classifier and $c$ is an output class, we have $c \in [\eta(x)\cup\mu(x)]_{G}$. Suppose $ c' \in [\eta(x)\cup\mu(x)]_{G}$. So $c'\in R(c)$ and by hypothesis we have $\mm, c' \models p_c$, which means that $(x,c') \in V(c) $, which implies that $c=c'$.

\noindent
If-part.
Suppose $[\eta(x)\cup\mu(x)]_{G} = \{c\}$. Because $c$ is an output of the classifier, $c \in W$, and as it is the only output of classifier $G$ on input $x$, we have $R(c) = \{c\}$. So it suffices to show that $\mm, c \models p_c$. As $\{(x,c)\} \in V(c)$, the statement is proved.  
\end{proof}
\begin{thm}
Suppose that  $\mathbb{G} = \{G_1, \dotsc, G_n\}$, is a multi-classifier system such that $\bigcap_{G\in \mathbb{G}}[\eta(x)\cup\mu(x)]_{G} \neq \emptyset$,  $x$ is an input point and $c$ is an output class for classifiers in $\mathbb{G}$, then
\begin{center}
    $\bigcap_{G\in \mathbb{G}}[\eta(x)\cup\mu(x)]_{G} = \{c\}$ if and only if $\mm, \Bar{c} \vDash \D{\mathbb{G}}p_c$,  where $p_c$ is interpreted by $(x,c)$.
\end{center}
\end{thm}
\begin{proof}
Only-if part.
 Suppose that $\mm, \Bar{c} \vDash \D{\mathbb{G}}p_c$, then for all $c'$ in $R_{D_{\mathbb{G}}}(\Bar{c})$ we have $\mm, c' \vDash p_c$. On the other hand, by the assumption there is an element $d$ in $\bigcap_{G\in \mathbb{G}}[\eta(x)\cup\mu(x)]_{G}$. It is enough to show that $d=c$. We have $d \in \bigcap_{G\in \mathbb{G}} R_{G_i}(\Bar{c})$. Since $\mm, d \vDash p_c$, then $V(d)=\{(x,c)\}$ which implies that $d=c$.
 
\noindent
If-part.
Suppose $\bigcap_{G\in \mathbb{G}}[\eta(x)\cup\mu(x)]_{G} = \{c\}$. Because $c$ is an output of the classifiers, $c \in W$, and $c \in [\eta(x)\cup\mu(x)]_{G}$, for all $G \in \mathbb{G}$. On the other hand as it is the only common output of all the classifiers, for any other output class, there exist a classifier $G$, that does not have it as output. So   $R_{\D{\mathbb{G}}} (c) =\{c, \Bar{c}\}$. Therefore it suffices to show that $\mm, \Bar{c} \models p_c$. As $\{(x,c)\} \in V(c)\cap V(\Bar{c})$, the statement is proved.  
\end{proof}


it is noteworthy that for each input point $x_0$, the possible worlds shall be representing all epistemic possible states (i.e., they show all possible subsets of output class set). These possible worlds can be filtered through the neighbourhood and manipulation set $\mu(x_0) \cup \eta(x_0)$. 
Let $C$ be the set of all output classes. If $|\mu(x_0) \cup \eta(x_0)| \geq |C|$, the number of all possible worlds shall be $2^{|C|}$ (generally $|\mu(x_0) \cup \eta(x_0)| \gg |C|$ can be assumed). Therefore, the satisfaction of an atomic formula $(x_0,c_i)$ in a state $w$ means that world $c_i$ has appeared as an output class of the classifier $G_k$, for some point of $\mu(x_0) \cup \eta(x_0)$. For instance, if $c_i$, $c_j$, and $c_k$ appear in the output classes of $G_k$, for all members of $\mu(x_0) \cup \eta(x_0)$, then the \textit{actual world} \cite{sep-possible-worlds} can satisfy $(x_0,c_i)$, $(x_0,c_j)$, and $(x_0,c_k)$. 
 \subsection{2-Multi-classifiers Systems}
\label{sec:2MCS}
In the DEL, the interpretation of external knowledge is also considered feasible. This means that in a MAS, besides the internal distributed knowledge, outer knowledge sharing can be investigated, in order to reduce the possible worlds. For example, in an optical character recognition (OCR) problem limited to numbers, knowledge of the existence of a circle in the shape of the input image could reduce to four possible worlds (0, 6, 8, and 9, which have at least a loop in their shape). Another benefit of considering external knowledge is the ability to divide a problem into smaller parts, and also to solve more simplistic problems using various MAS and the sharing of knowledge to conquer the problem. For this kind of divide-and-conquer algorithm in the MAS scenario, primarily, the problem should be broken down into simpler parts and the rule of division should be collected (here, rules are the restriction applied to the process of combination of the divisions), where each division can be solved with a MAS of classifiers. Secondly, all MASs should publicly announce their remaining possible worlds as external knowledge. Next, considering the rules, impossible combinations should be avoided to reduce the number of possible worlds. Finally, when one and only one possible world exists, it can be verified deterministically. For example, assume two classifiers that are supposed to classify an input image, one of which classifies the background and another that does the same for the foreground. Next, suppose that the first classifier has identified the background as a city street, and the second is uncertain regarding the foreground between the existence of an elephant or a car.  Using external knowledge that implies, ``elephants do not wander on city-streets'', the only possible world for the foreground would be: there is a \textit{car} on the street. 

Formally, Assume that 
$$\mathcal{M}_k = (W_k,R_{k,1},\cdots R_{k,n_k}, V_k),$$ where $1\leq k\leq n$,
are given. The Kripke model:\\
$$\mathcal{M}=(W,R_1, \cdots , R_n, V)$$
that models knowledge-sharing between classifiers, is defined through the following components:
\begin{itemize}
\item 
$W = W_1 \times \cdots\times W_n$, 
\item
$(w_{i_1},\cdots,w_{i_n}) R_k (w_{j_1},\cdots,w_{j_n})$, for $1\leq k \leq n$,
exactly when
for all $l \neq k$ we have $ w_{i_l}=w_{j_l}$ and  there exists ${1 \leq m \leq n_k}$ such that $ w_{i_k} R_{k,m} w_{j_k}$,
\item
$V(w_1, \cdots, w_n)= (V_1(w_1), \cdots , V_n(w_n))$.
\end{itemize}


\section{Artificial Neural Networks Collaboration}
\label{sec:Interactions}
Suppose that a group of trusted classifiers work together towards achieving a more aware system (where trusted classifiers share the entire owned knowledge correctly). Algorithm \ref{alg:CKC} would take a classifier, an input point, and a neighbourhood and manipulation function. This algorithm calculates all possible worlds for the classifier according to the input and its neighbourhoods and manipulations. As mentioned before, knowledge will be generated using this function. In other words, first of all, the process of knowledge extraction from one agent-input is developed, i.e., we collect all output classes of inputs in $\eta(x)$ related to the given classifier. In this algorithm, a classifier is a function $\mathcal{N}(x)$ for input $x$, and the output result of the function represents the output class of $x$. Consequently, $\mathcal{K}$ represents the knowledge of $\mathcal{N}$ with the above-mentioned definitions. As a result, the output represents whether the investigated classifiers are robust, for the input $x$, or not. And it includes the possible answers of the set $\eta(x)$ from the agent's perspective.
\begin{algorithm}
Let $\mathcal{N}(x)=c$ be the function of the considered a classifier and $c$ be the result class\;
\caption{The Classifier Knowledge Calculator (CKC) function shall calculate the knowledge produced by a classifier for an input point, using neighbourhood function and manipulation function }
\label{alg:CKC}
\begin{algorithmic}[1]
\Function{CKC}{$\mathcal{N},x_0, \eta, \mu$}\\
	\Comment{$\mathcal{N},x_0, \eta, \mu$ are a classifier, an input point, neighbourhood function, and manipulation function respectively}
	\State $\mathcal{K} \gets  \emptyset$ 
	\ForAll{$x \in \eta(x_0) \cup \mu(x_0)$}
		\State $c \gets  \mathcal{N}(x)$ \Comment{$c$ represents the respective possible world}
		\If{$c \notin \mathcal{K}$}
			\State Add $c$ to $\mathcal{K}$ set
		\EndIf
	\EndFor
	\If{$|\mathcal{K}| = 1$}
		\State \textbf{return} 1, $\mathcal{K}$
	\EndIf
	\State \textbf{return} 0, $\mathcal{K}$
\EndFunction
\end{algorithmic}
\end{algorithm}

After obtaining the knowledge of each agent (i.e., ANN -as classifiers), algorithm \ref{alg:MASKA} has been developed to aggregate all the knowledge of these agents. This knowledge also demonstrates the possible worlds from the perspective of each agent. Herein, by determining the intersected knowledge of all classifiers in the MAS, the result of verification, and the aggregated knowledge from the group of agents can be measured. As an output, if the intersection results in one class, the input is considered verified. Otherwise, if more than one class exists in the intersection result, the input point cannot be verified. Although, the set of possible classes represents the possible verified outputs for the MAS. Finally, if an empty set returns as an output, inconsistency emerges in the MAS's agent knowledge, and the input, for that particular case, cannot be verified. 
\begin{algorithm}
\caption{The MAS Knowledge Aggregator (MASKA) function is going to aggregate the knowledge that is produced by a group of classifiers, for each input point using a neighbourhood function and manipulation function. }\label{alg:MASKA}
\begin{algorithmic}[1]
\Function{MASKA}{$\mathcal{N_S},x_0, \eta, \mu$}\\
    \Comment{$\mathcal{N_S},x_0, \eta, \mu$ are a group of classifiers, an input point, neighbourhood function, and manipulation function respectively}
	\State $\mathcal{K_S} \gets  set~of~all~output~classes$ 
	\ForAll{$\mathcal{N} \in \mathcal{N_S}$}
		\State $is\_Robust$, $\mathcal{K} \gets  $ CKC $(\mathcal{N},x_0, \eta, \mu)$ 
		\State $\mathcal{K_S} \gets \mathcal{K_S} \cap \mathcal{K} $ 
		\If{$\mathcal{K_S} = \emptyset$}
			\State \textbf{return} 0, $\emptyset$
		\EndIf
	\EndFor
	\If{$|\mathcal{K_S}| = 1$}
		\State \textbf{return} 1, $\mathcal{K_S}$
	\EndIf
	\State \textbf{return} 0, $\mathcal{K_S}$
\EndFunction
\end{algorithmic}
\end{algorithm}

Algorithm \ref{alg:MASKS} would aggregate all classifiers' knowledge and external knowledge. 
Here, $\mathcal{K_S}$ shows the remaining possible worlds in which verification formulas can be satisfied if the cardinality of the remaining possible worlds equals one. In other words, $\mid \mathcal{K_S} \mid$ represents the number of words connected to $\Bar{w}$ with relations $R_i$, $i \in \mathcal{N_S}$.

\begin{algorithm}
\caption{The MAS Knowledge Sharing (MASKS) function shall aggregate the knowledge that is produced by an external system }\label{alg:MASKS}
\begin{algorithmic}[1]
\Function{MASKS}{$\mathcal{N_S},x_0, \eta, \mu$}\\
    \Comment{$\mathcal{N_S},x_0, \eta, \mu$ are a group of classifiers, an input point, neighbourhood function, and manipulation function respectively}
	\State $\mathcal{K_S} \gets  \emptyset$ 
	\ForAll{$\mathcal{N} \in \mathcal{N_S}$}
		\State $is\_Robust$, $\mathcal{K} \gets  $CKC$(\mathcal{N},x_0, \eta, \mu)$ 
		\State $\mathcal{K_S} \gets \mathcal{K_S} \cap \mathcal{K} $ 
		\If{$\mathcal{K_S} = \emptyset$}
			\State \textbf{return} 0, $\emptyset$
		\EndIf 
	\EndFor
	\If{$|\mathcal{K_S}| = 1$} \\ \Comment{Check whether the knowledge can verify the input, or if more knowledge is needed}
		\State \textbf{return} 1, $\mathcal{K_S}$
	\EndIf
	\ForAll{$\mathcal{M} \in $ All knowledge sources}
		\State $is\_Robust$, $\mathcal{K} \gets  $ Announced knowledge\\
		\Comment{The external knowledge must be written in the DEL formula, where possible worlds are ones in which the formula is satisfied.}
		\State $\mathcal{K_S} \gets \mathcal{K_S} \cap \mathcal{K} $ 
		\If{$\mathcal{K_S} = \emptyset$}
			\State \textbf{return} 0, $\emptyset$
		\EndIf
	\EndFor
	\If{$|\mathcal{K_S}| = 1$}
		\State \textbf{return} 1, $\mathcal{K_S}$
	\EndIf
	\State \textbf{return} 0, $\mathcal{K_S}$
\EndFunction
\end{algorithmic}
\end{algorithm}

\section{Examples}
\label{sec:EarlyResults}
Reducing the error rate of classifiers is crucial in critical scenarios. In our method, a MAS of classifiers has been offered for this exact purpose.
In the sequel, we will provide two types of examples explaining the details of our proposed model. In the first example, we demonstrate exactly how the knowledge set, in the context of dynamic epistemic logic, can be defined. In the other examples, we demonstrate the usability of our model in real-world practical applications. To achieve this, we select three datasets to examine the effects. The selected datasets are Modified-Fashion-MNIST \cite{xiao2017fashion} (unbalanced 4 output class), MNIST \cite{lecun1998mnist} (balanced 10 output class), and Fruits-360 \cite{murecsan2017fruit} (unbalanced 131 output class). A tool is also developed in order to apply the method in any arbitrary classifiers and datasets.

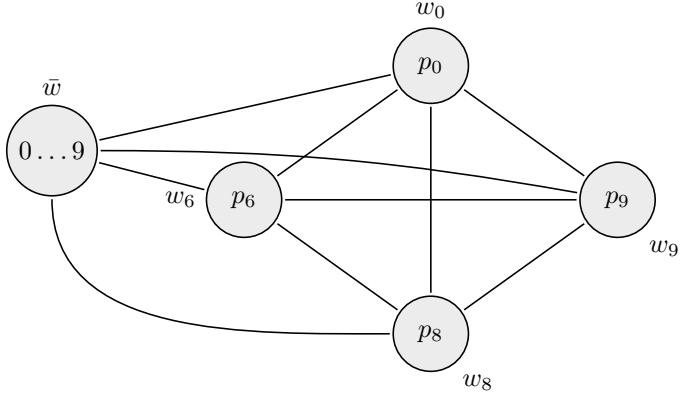
\begin{figure}       
\begin{tikzpicture}[modal]
\node[world,minimum size= 1.0cm] (w0) [label=above:{$w_0$},] { $p_0$  };
\node[world,minimum size= 1.0cm] (w6) [label=180:{$w_6$},below left =of w0, xshift=-7mm, yshift=0mm] {$p_6$ };
\node[world,minimum size= 1.0cm] (w) [label=above:{$\Bar{w}$},above left =of w6, xshift=-7mm, yshift=-12mm] { $0 \dots 9$ };
\node[world,minimum size= 1.0cm] (w8) [label=305:{$w_8$},below right =of w6, xshift=7mm, yshift=0mm] {$p_8$  };
\node[world,minimum size= 1.0cm] (w9) [label=305:{$w_9$},above right =of w8, xshift=7mm, yshift=0mm] {$p_9$  };
\path[-] (w) edge (w6)[color=black];
\path[-] (w) edge[out=270,in=180] (w8)[color=black];
\path[-] (w) edge[out=0,in=170] (w9)[color=black];
\path[-] (w) edge (w0)[color=black];
\path[-] (w0) edge (w6)[color=black];
\path[-] (w0) edge (w8)[color=black];
\path[-] (w0) edge (w9)[color=black];
\path[-] (w6) edge (w8)[color=black];
\path[-] (w9) edge (w8)[color=black];
\path[-] (w6) edge (w9)[color=black];
\end{tikzpicture}
  \caption{The Epistemic model, considering the classifier $A_0$'s knowledge. \newline 
$W^{(p_0 \lor p_6 \lor p_8 \lor p_9)} =\{w_0,w_6,w_8,w_9, \Bar{w}\}$;\newline 
$R_{A_0}=\{(w_0,w_0),(w_6,w_6),(w_8,w_8),(w_0,w_8),(w_8,w_0),(w_0,w_9),(w_9,w_0),\\(w_6,w_8),(w_8,w_6),(w_6,w_9),(w_9,w_6),(w_8,w_9),(w_9,w_8), (\Bar{w}, \Bar{w}), (\Bar{w}, w_0),\\(w_0,\Bar{w}), (\Bar{w}, w_6),(w_6,\Bar{w}), (\Bar{w}, w_8),(w_8,\Bar{w}), (\Bar{w}, w_9),(w_9,\Bar{w})\}$; }
  \label{fig:figA0}
  \end{figure}

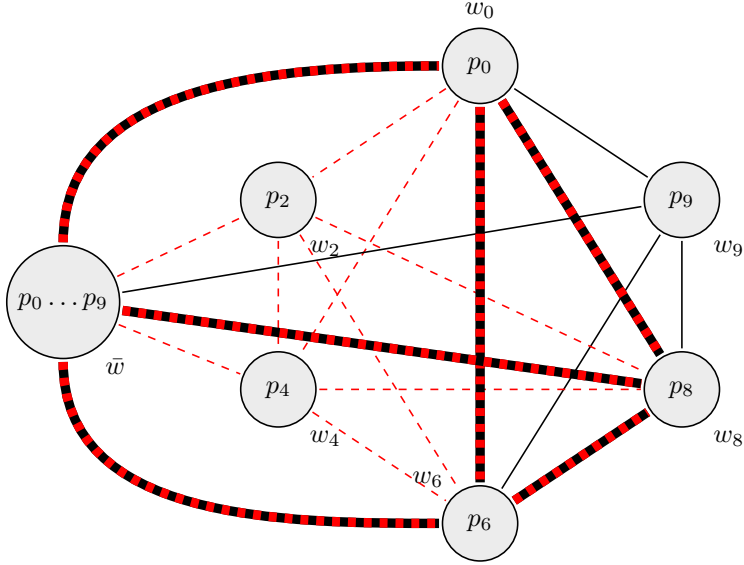
\begin{figure}
\begin{tikzpicture}[modal]
\node[world,minimum size= 1.0cm] (w0) [label=above:{$w_0$},] { $p_0$  };
\node[world,minimum size= 1.0cm] (w2) [label=305:{$w_2$},below left =of w0, xshift=-9mm, yshift=0mm] {$p_2$  };
\node[world,minimum size= 1.0cm] (w4) [label=305:{$w_4$},below =of w2, xshift=0mm, yshift=0mm] {$p_4$  };
\node[world,minimum size= 1.0cm] (w6) [label=135:{$w_6$},below right =of w4, xshift=9mm, yshift=0mm] {$p_6$ };
\node[world,minimum size= 1.0cm] (w8) [label=305:{$w_8$},above right =of w6, xshift=9mm, yshift=0mm] {$p_8$  };
\node[world,minimum size= 1.0cm] (w9) [label=305:{$w_9$},above =of w8, xshift=0mm, yshift=0mm] {$p_9$ };
\node[world,minimum size= 1.0cm] (w) [label=305:{$\Bar{w}$},below left =of w2, xshift=-9mm, yshift=6mm] {$p_0 \dots p_9$  };
\path[-] (w0) edge (w9)[color=black];
\path[-] (w9) edge (w8)[color=black];
\path[-] (w6) edge (w9)[color=black];

\path[dashed] (w0) edge (w2)[color=red];
\path[dashed] (w0) edge (w4)[color=red];

\path[dashed] (w2) edge (w4)[color=red];
\path[dashed] (w2) edge (w6)[color=red];
\path[dashed] (w2) edge (w8)[color=red];

\path[dashed] (w4) edge (w6)[color=red];
\path[dashed] (w4) edge (w8)[color=red];

\path[dashed] (w) edge (w2)[color=red];
\path[dashed] (w) edge (w4)[color=red];
\path[-] (w) edge (w9)[color=black];
\path[line width=1.25mm,-] (w0) edge (w6)[color=black];
\path[line width=1.25mm,dashed] (w0) edge (w6)[color=red];
\path[line width=1.25mm,-] (w0) edge (w8)[color=black];
\path[line width=1.25mm,dashed] (w0) edge (w8)[color=red];
\path[line width=1.25mm,-] (w6) edge (w8)[color=black];
\path[line width=1.25mm,dashed] (w6) edge (w8)[color=red];
\path[line width=1.25mm,-] (w) edge[out=270,in=180] (w6)[color=black];
\path[line width=1.25mm,dashed] (w) edge[out=270,in=180] (w6)[color=red];
\path[line width=1.25mm,-] (w) edge (w8)[color=black];
\path[line width=1.25mm,dashed] (w) edge (w8)[color=red];
\path[line width=1.25mm,-] (w) edge[out=90,in=180] (w0)[color=black];
\path[line width=1.25mm,dashed] (w) edge[out=90,in=180] (w0)[color=red];
\end{tikzpicture}
  \caption{The updated model, after $A_1$'s announcement. The red-black relations are intersected relations of the model, the possible worlds after the announcement will be $w_0,~ w_6,~ and~ w_8$  \newline 
$W^{(p_0 \lor p_6 \lor p_8 \lor p_9) \land (p_0 \lor p_2 \lor p_4 \lor p_6 \lor p_8)}=\{w_0,w_6,w_8, \Bar{w}\}$;\newline 
$R_{A_0} \cap R_{A_1}=\{(w_0,w_0),(w_6,w_6),(w_8,w_8),(w_0,w_6),(w_6,w_0),(w_0,w_8),(w_8,w_0),\\(w_6,w_8),(w_8,w_6), (\Bar{w}, \Bar{w}), (\Bar{w}, w_0), (w_0, \Bar{w}), (\Bar{w}, w_6), (w_6, \Bar{w}), (w_8, \Bar{w}), (\Bar{w}, w_8)\}$; }
  \label{fig:figA0A1}
  \end{figure}

\begin{exa}(Digit Recognition)
\label{exam1}
\end{exa}
In this example, we will develop a scenario to clarify the approach mentioned in previous chapters. Herein, the input will lie in the domain of images. One of the digits 0 to 9 would appear in the image. The model has eleven worlds $W=\{ w_0,\cdots,w_9 \} \cup \Bar{w}$. Each world $w_i$ represents the existence of one digit and $\Bar{w}$, where   $V_\Phi(w_i)=\{(x,i)\}$. The world $\Bar{w}$ could not be a possible answer because more than one digit is true in it. This world was added to check the satisfaction of the verification formula (as mentioned in previous sections) $V_\Phi(\Bar{w})=\{(x,0), \dots, (x,9)\}$. For a fixed input of $x$, we denote the atomic formula $(x, i)$ by $p_i$.
Let the figure of the digit ``0'' be an input image, and let a multi-agent system $A$ with three given agents $A = \{A_0, A_1,  A_2\}$ as classifiers. Assume that $A_0$ shows ``0'', ``6'', ``8'', and ``9'' are possible answers for the input $x_0$. Figure \ref{fig:figA0} depicts the possible worlds and relations of the model for $A_0$.
The same input for $A_1$ results ``0'', ``2'', ``4'', ``6'', and ``8'' are possible answers. The intersection of these two agents' possible answers will be $w_0$, $w_6$, and $w_8$, fig \ref{fig:figA0A1}. The last agent, $A_2$, possible results are $w_0$, $w_2$, and $w_9$. Thus it ignores the world $w_6$ and $w_8$ as possible results. So, the model verifies the properties for the answer that the input case is ``0'' \ref{fig:figA0A1A2}. We have $\mm, \Bar{w} \vDash \D{A}p_0$. 


  \begin{figure}
\begin{tikzpicture}[modal]
\node[world,minimum size= 1.0cm] (w0) [label=above:{$w_0$},] { $p_0$  };
\node[world,minimum size= 1.0cm] (w2) [label=305:{$w_2$},below left =of w0, xshift=-9mm, yshift=0mm] {$p_2$  };
\node[world,minimum size= 1.0cm] (w4) [label=305:{$w_4$},below =of w2, xshift=0mm, yshift=0mm] {$p_4$  };
\node[world,minimum size= 1.0cm] (w6) [label=135:{$w_6$},below right =of w4, xshift=9mm, yshift=0mm] {$p_6$ };
\node[world,minimum size= 1.0cm] (w8) [label=305:{$w_8$},above right =of w6, xshift=9mm, yshift=0mm] {$p_8$  };
\node[world,minimum size= 1.0cm] (w9) [label=305:{$w_9$},above =of w8, xshift=0mm, yshift=0mm] {$p_9$ };
\node[world,minimum size= 1.0cm] (w) [label=305:{$\Bar{w}$},below left =of w2, xshift=-9mm, yshift=6mm] {$p_0 \dots p_9$  };
\path[-] (w0) edge (w6)[color=black];
\path[-] (w0) edge (w8)[color=black];
\path[-] (w0) edge (w9)[color=black];
\path[-] (w9) edge (w8)[color=black];
\path[-] (w6) edge (w9)[color=black];
\path[-] (w6) edge (w8)[color=black];

\path[dashed] (w0) edge (w2)[color=red];
\path[dashed] (w0) edge (w4)[color=red];

\path[dashed] (w2) edge (w4)[color=red];
\path[dashed] (w2) edge (w6)[color=red];
\path[dashed] (w2) edge (w8)[color=red];

\path[dashed] (w4) edge (w6)[color=red];
\path[dashed] (w4) edge (w8)[color=red];
\path (w0) edge[zigzag, out=200,in=55] (w2)[color=blue];
\path (w0) edge[zigzag, out=0,in=90] (w9)[color=blue];
\path (w2) edge[zigzag] (w9)[color=blue];

\path[-] (w) edge[out=45,in=190] (w0)[color=black];
\path[-] (w) edge (w9)[color=black];
\path[-] (w) edge[out=310,in=180] (w8)[color=black];
\path[-] (w) edge[bend right] (w6)[color=black];
\path[dashed] (w) edge (w2)[color=red];
\path[dashed] (w) edge (w4)[color=red];
\path[dashed] (w) edge[out=305,in=160] (w6)[color=red];
\path[dashed] (w) edge (w8)[color=red];
\path[line width=1.25mm,-] (w) edge[bend left] (w0)[color=black];
\path[line width=1.25mm,dashed] (w) edge[bend left] (w0)[color=red];
\path (w) edge[zigzag, bend left] (w0)[color=blue];
\end{tikzpicture}
  \caption{The updated model, after $A_2$'s announcement. The red-black with blue zigzag relation is intersected relations of the model, $w_0$ would be the possible world after the announcement \newline 
$W^{(p_0 \lor p_6 \lor p_8 \lor p_9) \land (p_0 \lor p_2 \lor p_4 \lor p_6 \lor p_8)\land (p_0 \lor p_2 \lor p_9 )}=\{w_0, \Bar{w}\}$;\newline 
$R_{A_0} \cap R_{A_1} \cap R_{A_2} =\{(w_0,w_0), (\Bar{w}, \Bar{w}), (\Bar{w}, w_0), (w_0, \Bar{w})\}$; }
  \label{fig:figA0A1A2}
  \end{figure}

\begin{figure}[!ht]
\centering
\includegraphics[width=0.85\textwidth]{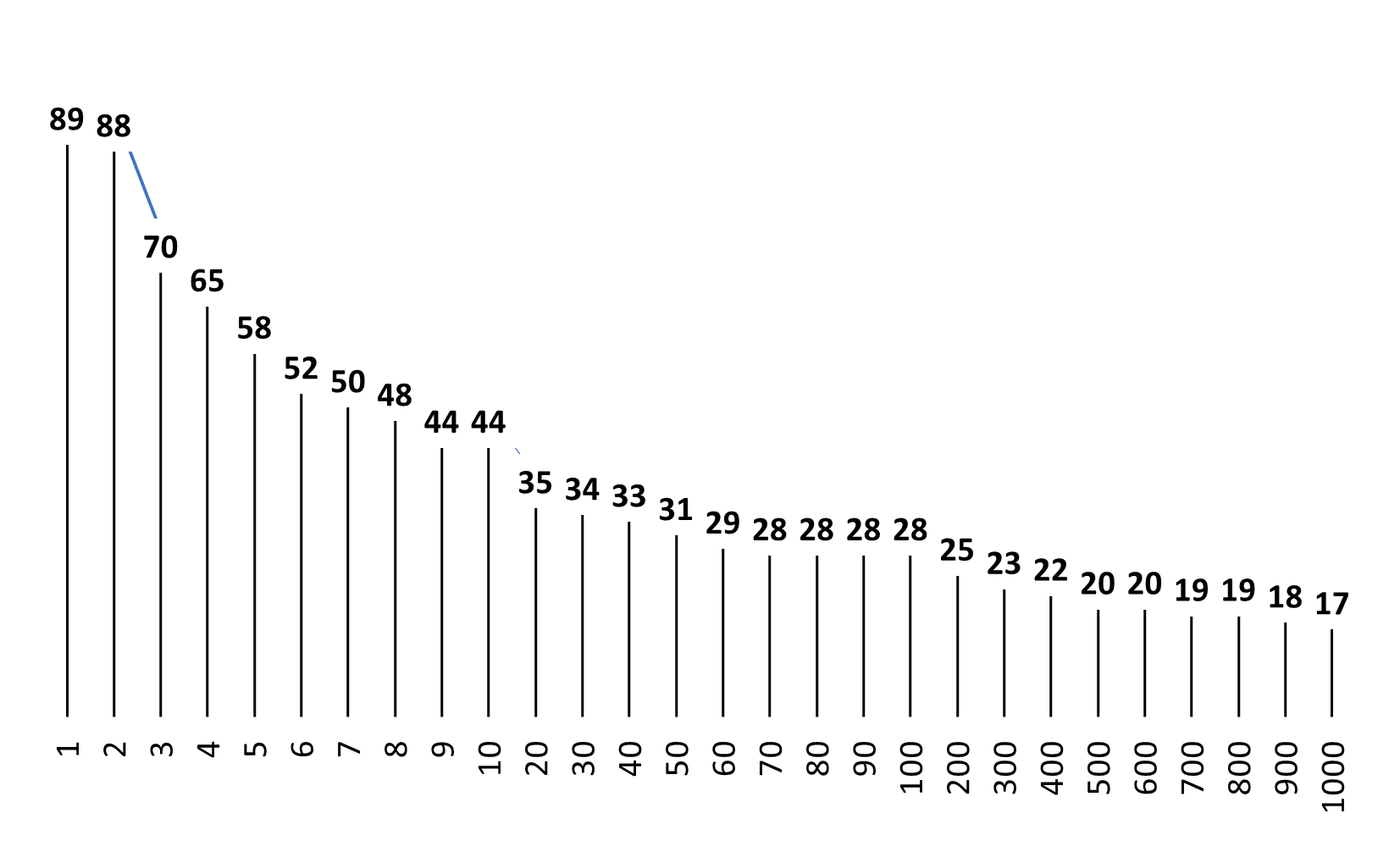}
\caption{Fashion-MNIST: wrong verified answers for various number of classifiers}
\label{fig:Fashion-MNIST-Err}
\end{figure}

\begin{exa}(Modified Fashion MNIST: The unbalanced-4-output-class example)
\label{exam2} 
\end{exa}
The Fashion-MNIST is a dataset of clothes' images that contains a training set of 60,000 examples and a test set of 10,000. Each image is a 28 by 28 matrix, associated with a label from ten classes. In this research, we modified the ten output classes of Fashion-MNIST (``T-shirt/top'', ``Trouser'', ``Pullover'', ``Dress'', ``Coat'', ``Sandal'', ``Shirt'', ``Sneaker'', ``Bag'', and ``Ankle boot'') to four ( ``Top cover'', ``Bottom cover'', ``Shoes'', and ``Bag'' in which ``Top cover'' contains ``T-shirt/top'', ``Pullover'', ``Dress'', ``Coat'', and ``Shirt''; ``Bottom cover'' contains ``Trouser''; ``Shoes'' contains ``Sandal'', ``Sneaker'', and ``Ankle boot''; and ``Bag'' contains ``Bag'';) in order to reach an unbalanced dataset with four output classes. Here, the neighbourhood and manipulation set (property to verify) is a set of 100 random salt-and-pepper noises (noises are created before the classification process and fixed for all inputs and classifiers). Although creating a more meaningful neighbourhood and manipulation set could boost our results, we decided to select it randomly in order to simulate errors in selecting neighbourhood and manipulation for real applications. In this example, all classifiers' accuracy is about 99 percent. This example could examine our method for data sets with few output classes. We execute the MASKS for a single to 1000 classifiers (see table \ref{tab:results}).  
{\color{black} Considering the neighbourhood and manipulation set, the number of wrong verified answers for a single classifier is 89, which means 89 out of 10000 cases are robust (or verified), but the presented answer is incorrect. These wrong verified answers occur when the classifier cannot correctly classify the input image and cannot find the correct output for the input's neighbourhood and manipulations. The value of wrong verified answers is close to the classifier's error value for the original input (without neighbourhoods and manipulations). Thus, in this case, the neighbourhood and manipulation set did not select ideally because all neighbourhoods and manipulation for the classifier result in a wrong answer. As mentioned before, choosing a neighbourhood and manipulation set is very complex. Here, wrong verified answers could be reduced by increasing the number of classifiers (with the same neighbourhood and manipulation set). In this example, increasing the number of classifiers could reduce the wrong verified answers caused mainly by improper neighbourhood and manipulation set selection.}
As we can see, the methods reduce the errors by about 80.5 percent. The detail of results for 1 to 1000 classifiers are represented in fig \ref{fig:Fashion-MNIST-Err}.

\begin{figure}[!ht]
\centering
\includegraphics[width=0.85\textwidth]{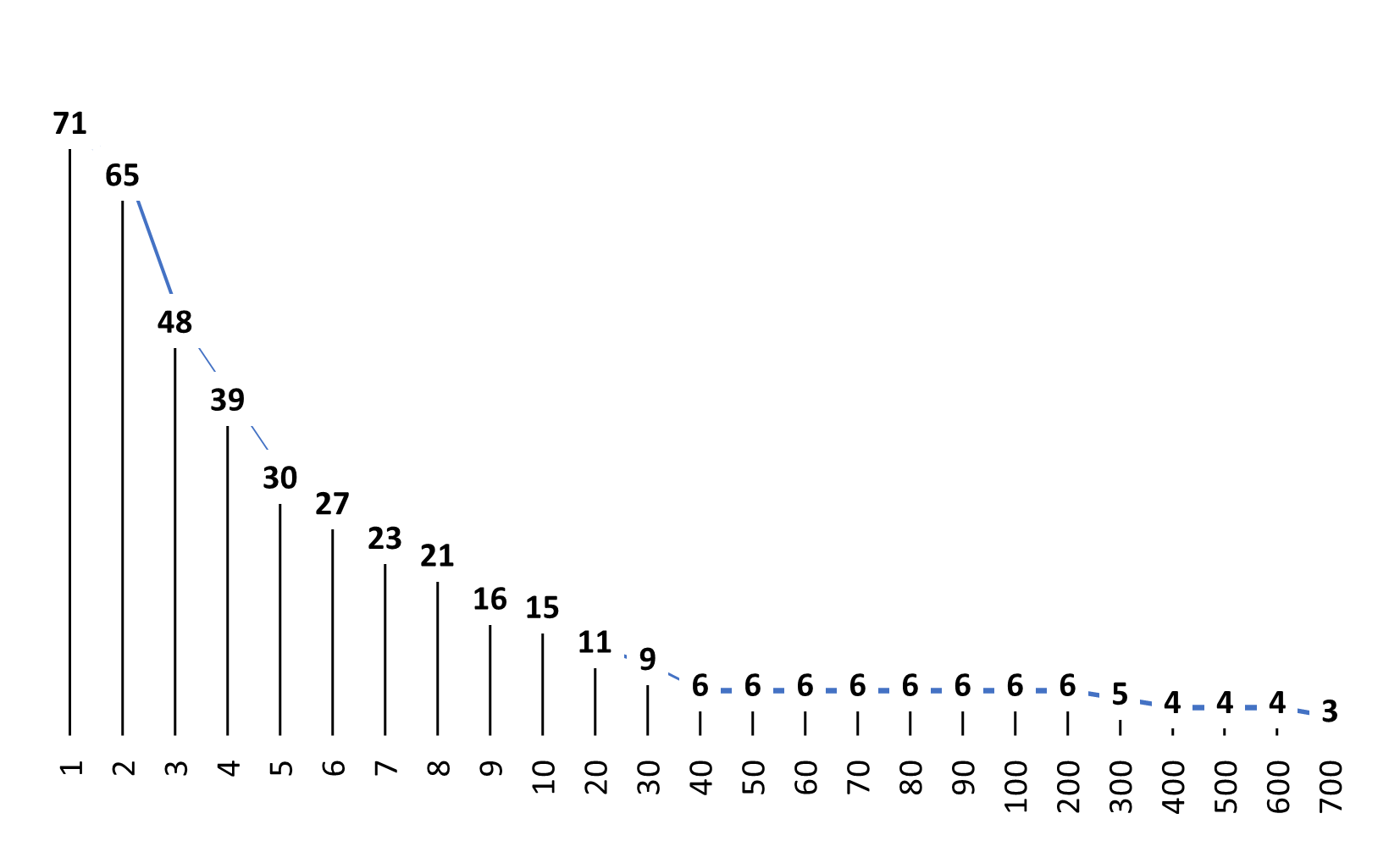}
\caption{MNIST: wrong verified answers for various number of classifiers}
\label{fig:MNIST-Err}
\end{figure}

\begin{exa}(MNIST: The unbalanced-10-output-class example)
\label{exam3} 
\end{exa}
The MNIST is a dataset of handwriting digit images that contains a training set of 60,000 examples and a test set of 10,000. Each image is a 28 by 28 matrix, associated with a label from ten classes. Here, the neighbourhood and manipulation set (property to verify) is a set of 100 noisy salt-and-pepper images (noises are created before the classification process and fixed for all inputs and classifiers); and the classifiers' accuracy is about 99\%. We execute the MASKS for a single to 608 agents (see table \ref{tab:results}).
{\color{black} In this example, the wrong verified answers for a single classifier are less than the previous example with the same accuracy. This example shows that the random noises are working better, but it is not ideal. Perhaps, increasing the number of output classes is a cause. In this example, increasing the number of classifiers could reduce wrong verified answers faster. As we can see here, the method reduces errors by about 95.75 percent with $\sim$ 700 classifiers. In this example, increasing the number of classifiers could reduce the wrong verified answers caused by the improper selection of neighbourhood and manipulation set and classifier's accuracy.
} 
The detailed results for 1 to 608 agents are represented in fig \ref{fig:MNIST-Err}.

\begin{figure}[!ht]
\centering
\includegraphics[width=0.85\textwidth]{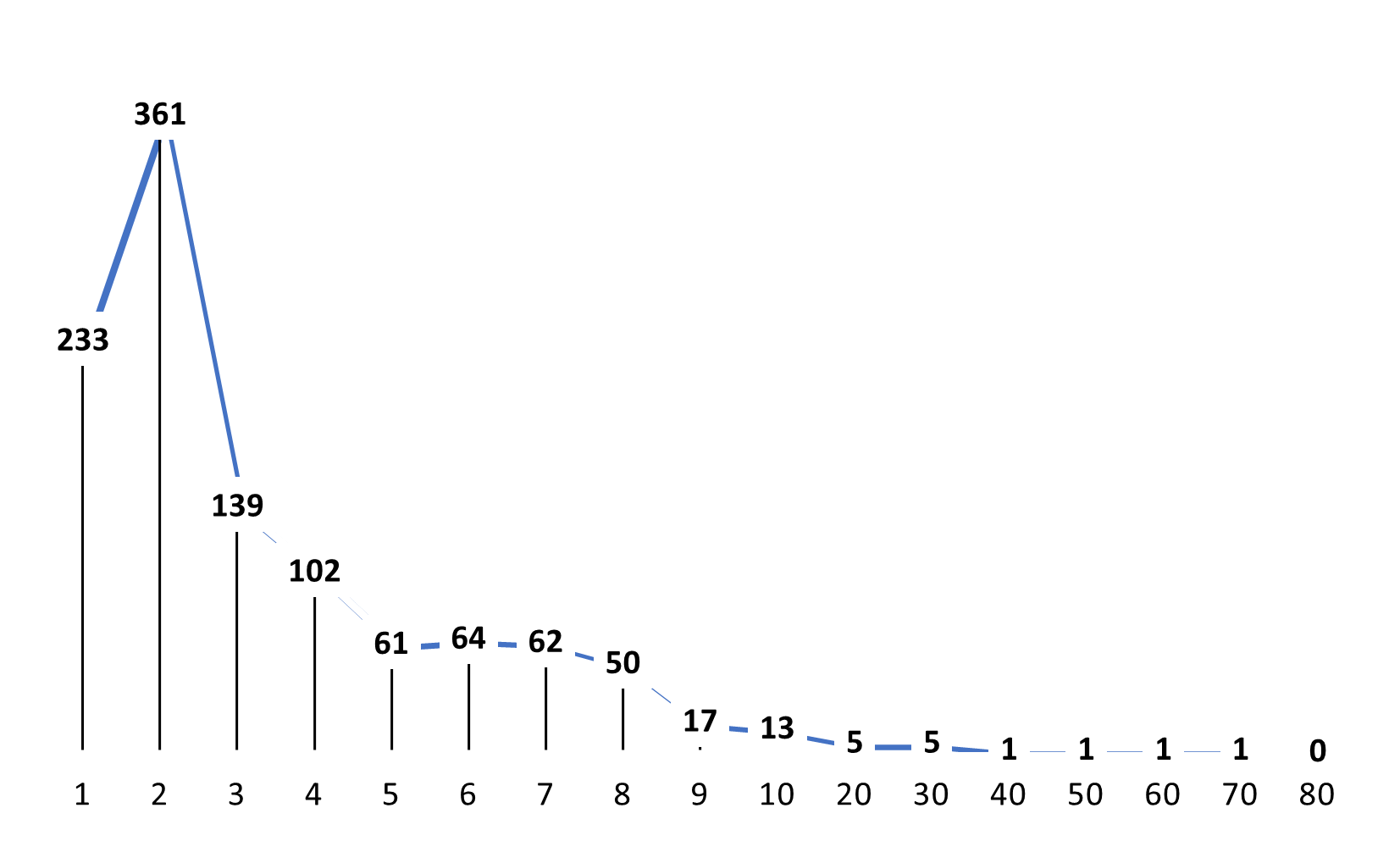}
\caption{Fruit-360: wrong verified answers for a various number of classifiers}
\label{fig:Fruit-360-Err}
\end{figure}

\begin{exa}(Fruit-360: The unbalanced-131-output-class example)
\label{exam4} 
\end{exa}
Fruits-360 is a dataset of fruits and vegetable images; the set contains 67692 training and 22688 test images (100 by 100) of 131 fruits and vegetables label classes.
Here, the neighbourhood and manipulation set is a set of 50 noisy salt-and-pepper images; and the classifiers' accuracy is about 97 percent. We execute the MASKS for single to 74 classifiers (see table \ref{tab:results}). 
{\color{black} In this example, the percentage of wrong verified answers of verified cases for a single classifier is even less than in the previous example. For a single classifier, wrong verified cases are about thirds of error cases (errors of original inputs without neighbourhoods and manipulations). It shows that the random noises are working better in this example. As we can see here, the number of wrong verified cases for two classifiers is more than a single classifier; this means two classifiers are agreed for more wrong verified answers. Thus, it seems the impact of accuracy is significant. The wrong verified cases reduce by increasing classifiers. 
}
As we can see here, all errors are disappeared in verified cases. The detailed results for 1 to 74 agents are represented in fig \ref{fig:Fruit-360-Err}.

\begin{rem}(Conclusion of examples)
\label{sec:ConclusionEX}
\end{rem}

\begin{figure}[!ht]
\centering
\includegraphics[width=0.85\textwidth]{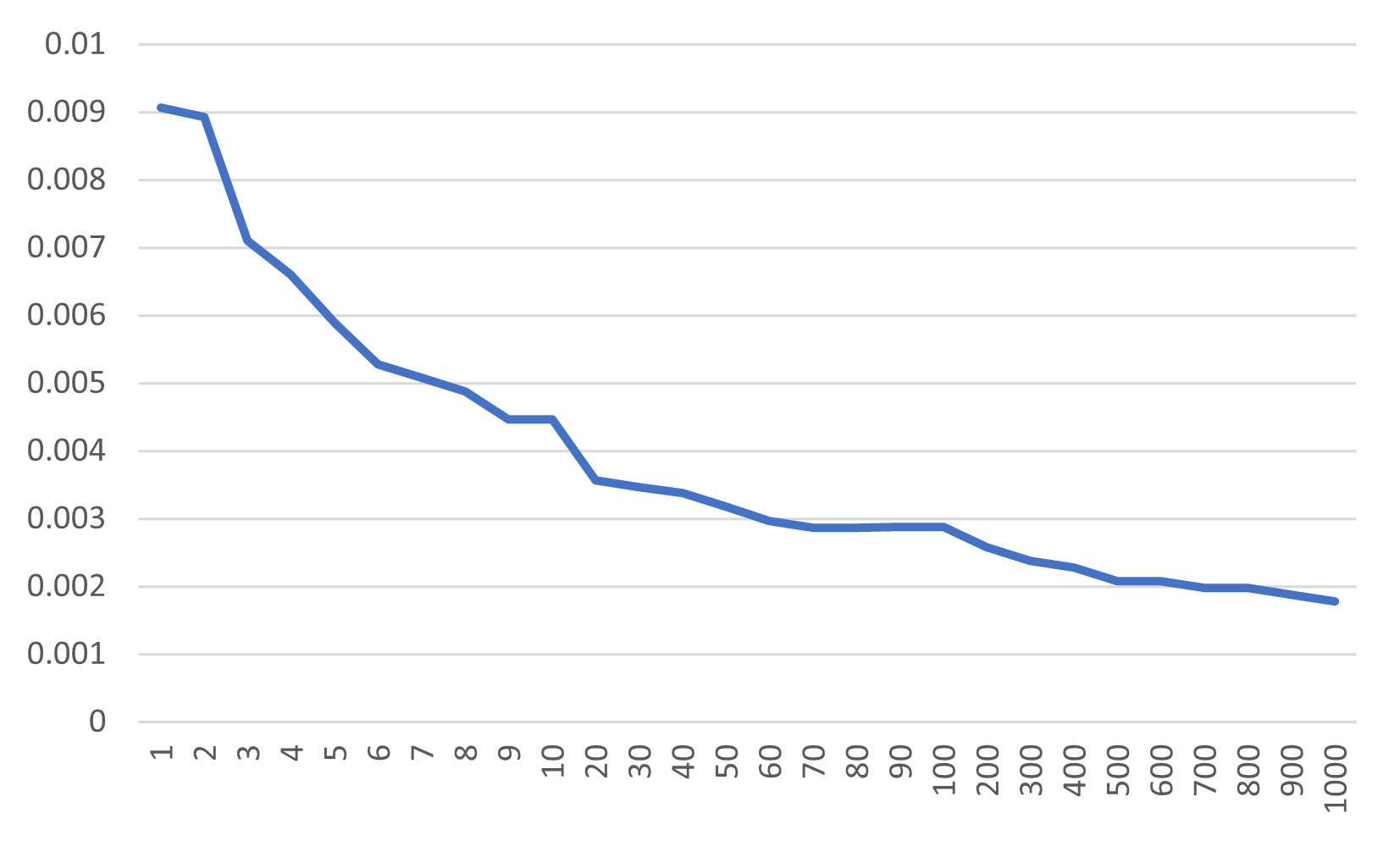}
\caption{Fashion-MNIST: ``wrong verified cases''/``wrong verified cases'' rate for various number of classifiers.}
\label{fig:FashionMNIST-WrongVerified-CorrectVerified}
\end{figure}
\begin{figure}[!ht]
\centering
\includegraphics[width=0.85\textwidth]{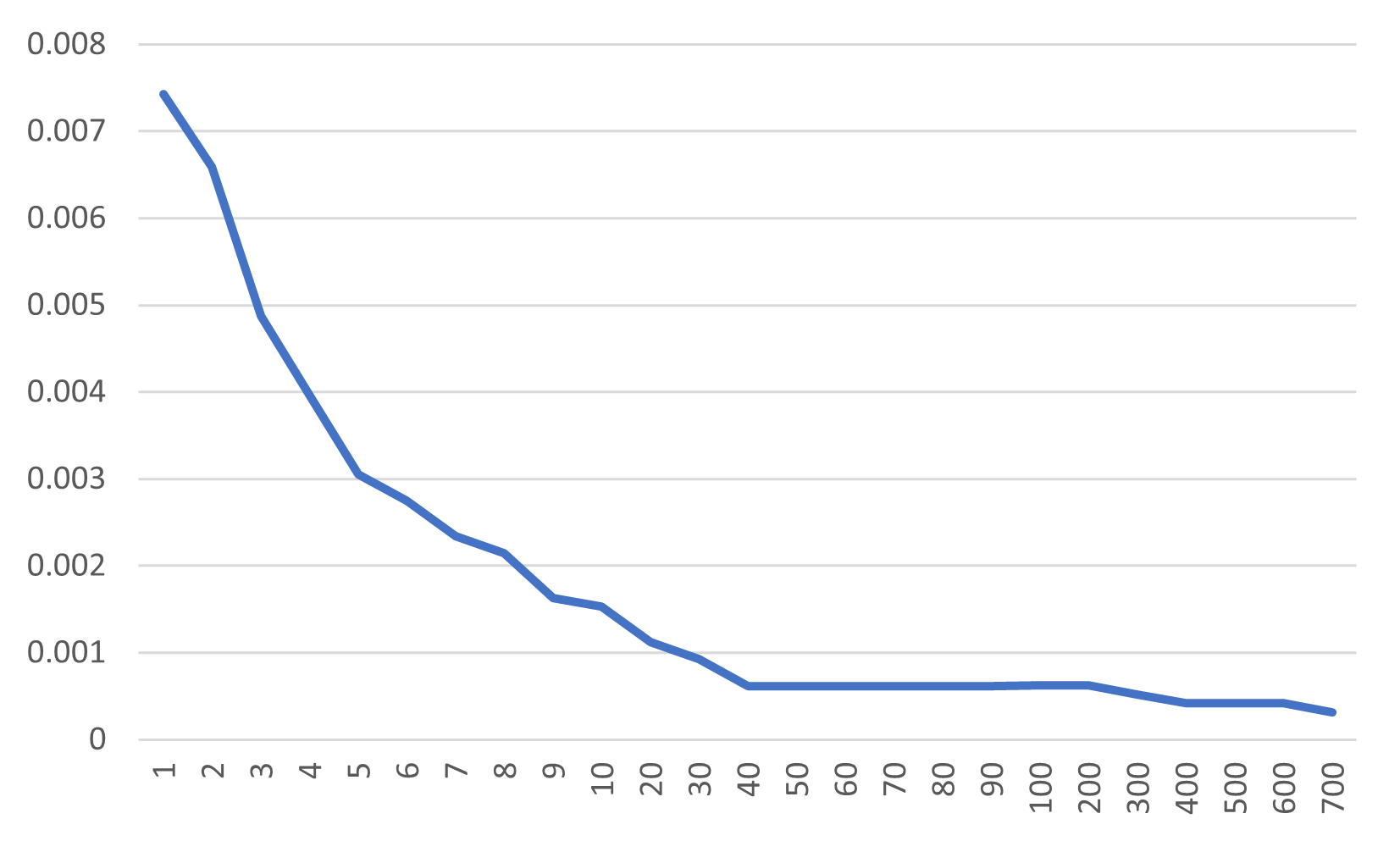}
\caption{MNIST: ``wrong verified cases''/``wrong verified cases'' rate for various number of classifiers.}
\label{fig:MNIST-WrongVerified-CorrectVerified}
\end{figure}
\begin{figure}[!ht]
\centering
\includegraphics[width=0.85\textwidth]{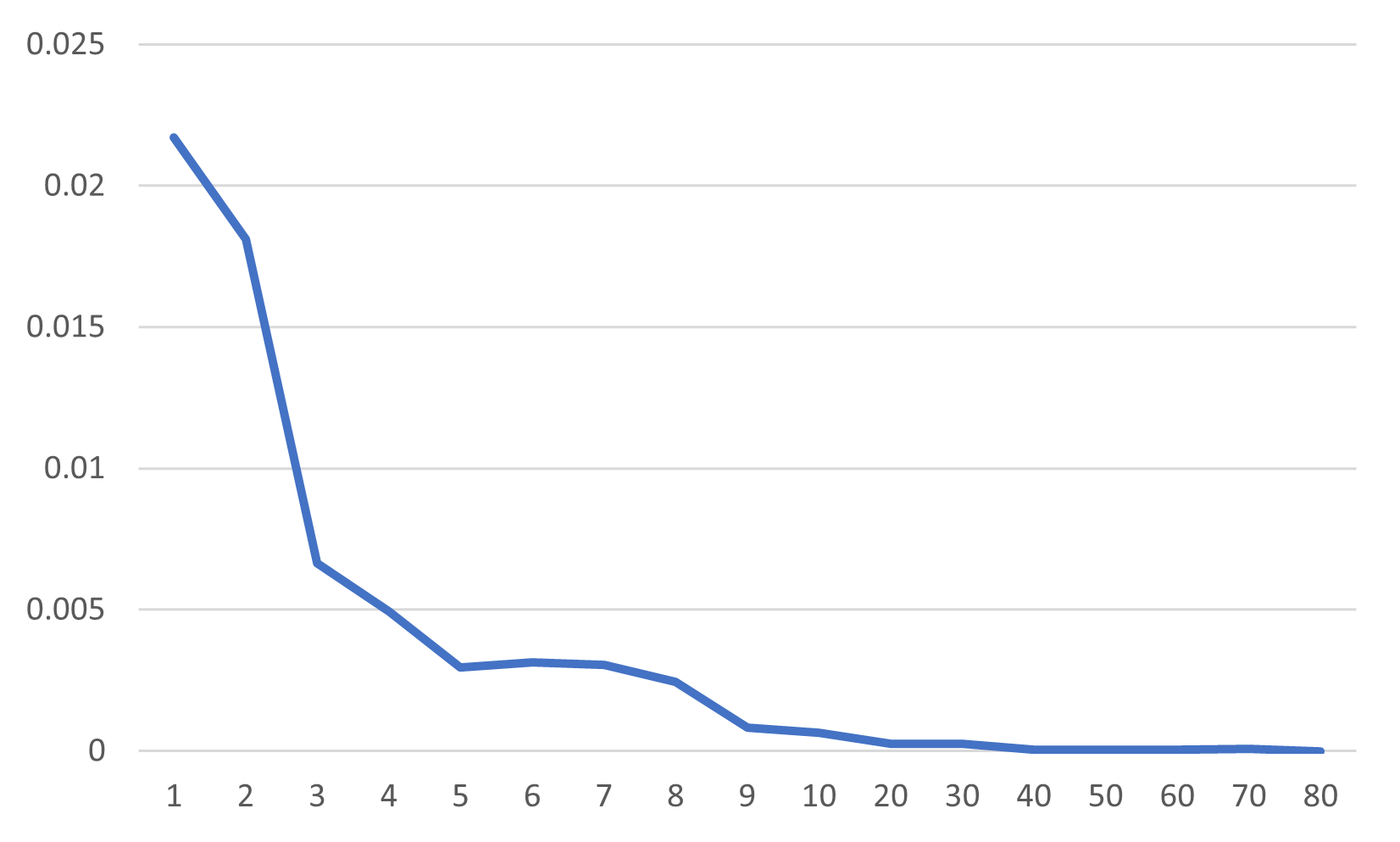}
\caption{Fruit-360: ``wrong verified cases''/``wrong verified cases'' rate for various number of classifiers.}
\label{fig:Fruit-360-WrongVerified-CorrectVerified}
\end{figure}

In these examples, a MAS of ANNs (as classifiers) has been developed to classify these datasets. As mentioned before, the input picture and all of the manipulation and the neighbourhood set will be fed into classifiers. If a single output class is represented as output, this input is verified (for the manipulation and the neighbourhood set). Otherwise, the algorithm does not represent any output for it. As is shown in fig \ref{fig:Fashion-MNIST-Err}, \ref{fig:MNIST-Err}, and \ref{fig:Fruit-360-Err}, for a system with one classifier, the number of verified input cases that are wrongly classified is much less than the multi-classifiers variant. These examples with one classifier are the same as the method developed in \cite{Huang0}, in which neighbourhood and manipulations are applied just in input (not in deeper layers). Compared with a single classifier,  it appears that an increased number of classifiers could reduce errors. Even though the number of correctly predicted cases also decreases with the increasing number of classifiers, fig \ref{fig:FashionMNIST-WrongVerified-CorrectVerified}, \ref{fig:MNIST-WrongVerified-CorrectVerified}, and \ref{fig:Fruit-360-WrongVerified-CorrectVerified} are ratios of ``wrong verified cases'' by ``correct verified cases'', which shows that the trend of decreasing the ``correct verified cases'' is not as steep as the slope of the ``wrong verified cases''.
Note that the model will decide for verified cases; thus, the predicted cases are verified ones. Other cases (unverified ones) did not count as ``correct predicted cases'' or ``wrong predicted cases''. 
Here, the unverified cases could be categorised as high-risk inputs. These inputs are the ones in which classifiers in the developed MAS do not show a good confidence level regarding the classification. 

As it can be seen, when we increase the number of classifiers, the error reduction is significant. Therefore, the method is well suitable for classifying critical cases (i.e., medical). In such cases, multiple classifiers could be applied to solve a classification problem with a small error ratio for verified cases. Unverified cases could be classified by a more trustworthy classifier (i.e., doctors in medical cases). Moreover, as it can be seen, defining a suitable property is complex and inaccurate. Thus, in these examples, we did not define a very complex property (random noise as neighbourhood and manipulation set) to show that even with an inaccurate property, MASKS could reduce errors.

\begin{rem}(Difficulties)
{\color{black} It is known that the verification methods goals are developed to satisfy specific predefined properties. However, the classification generally aims to reduce errors. The verification methods could help classifiers to be self-aware if properties are defined correctly. Nevertheless, defining such properties is difficult in many cases. For example, safety properties are applied in many image classification methods, but defining the $\epsilon$ for neighbourhood and the function $f$ in manipulation is not trivial. Examples \ref{exam2},\ref{exam3}, and \ref{exam4} will show how multi-classifiers could reduce the misclassification caused by improper safety properties.  }
Accordingly, our developed systems of classifiers may have miscalculation problems, if faced with the following scenarios: 
\begin{enumerate}
\item \textbf{Wrongly verifying input points:} This occurs when an input point has been considered as robust in a wrong class (i.e., the input point and all members of the knowledge set lie in the wrong output class). This failure may happen when the inaccuracy of the partitioning algorithm of the classifiers is more than the broadness of the neighbourhood and the manipulation set. The root cause of this can be in applying low classification accuracy, selecting an inadequately sized set for the neighbourhood and manipulation functions, or it can result from an insufficient number of classifiers employed in the MAS. 

\textbf{Note:} In the case that neighbourhood and manipulation set just contains the input point, this problem could be considered as ensemble learning. The probability of such error depends on every single classifier of the MAS (see Dietterich \textit{et al.} \cite{dietterich2000ensemble}).
\item \textbf{Correct answers not verified:} Sometimes, input points that are correctly classified with the classifiers are not verified by the MAS. In this scenario, the input point should be located near the partitioning boundaries of all the classifiers of the MAS. In other words, there exist similar inputs (which are elements of the neighbourhood and manipulation set) which have been wrongly classified into the same class, by all classifiers. In this case, at least two classes are presented as output class, by all classifiers. Through examining every intersection of output classes for the outputs of classifiers, a subset of classes can be collected. This subset is considered common among outputs for each classifier. Although the input point cannot be verified for a single result, it is acknowledged that the verified element resides in a subset of the represented results.  
\end{enumerate}
\end{rem}

\begin{figure}[!ht]
\centering
\includegraphics[width=0.85\textwidth]{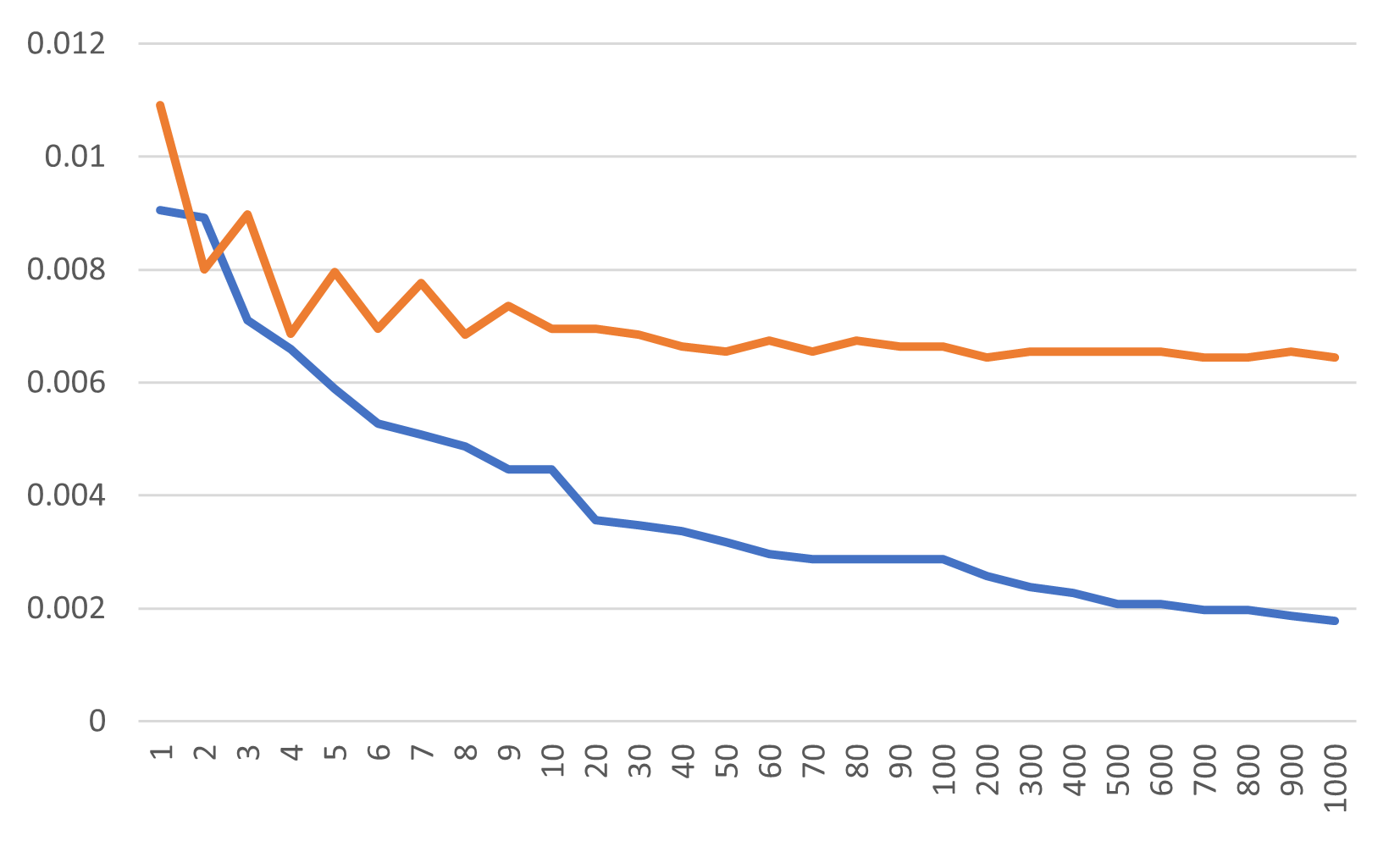}
\caption{Fashion-MNIST: Comparing wrong verified cases of various number of classifiers for MASKS and Voting.}
\label{fig:Fashion-MNIST-Err-vs-voting}
\end{figure}

\begin{figure}[!ht]
\centering
\includegraphics[width=0.85\textwidth]{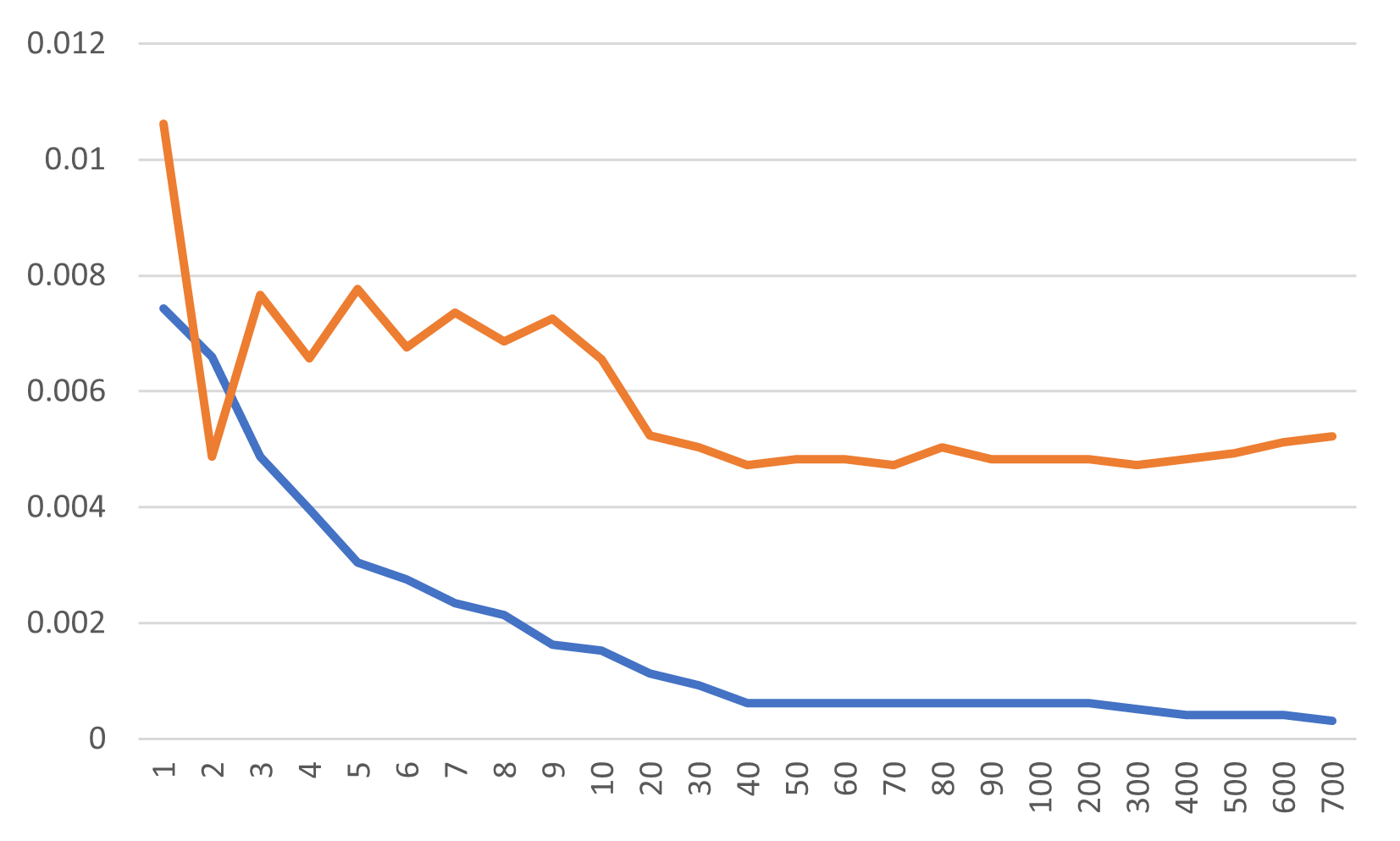}
\caption{MNIST: Comparing wrong verified cases of various number of classifiers for MASKS and Voting.}
\label{fig:MNIST-Err-vs-voting}
\end{figure}

\begin{figure}[!ht]
\centering
\includegraphics[width=0.85\textwidth]{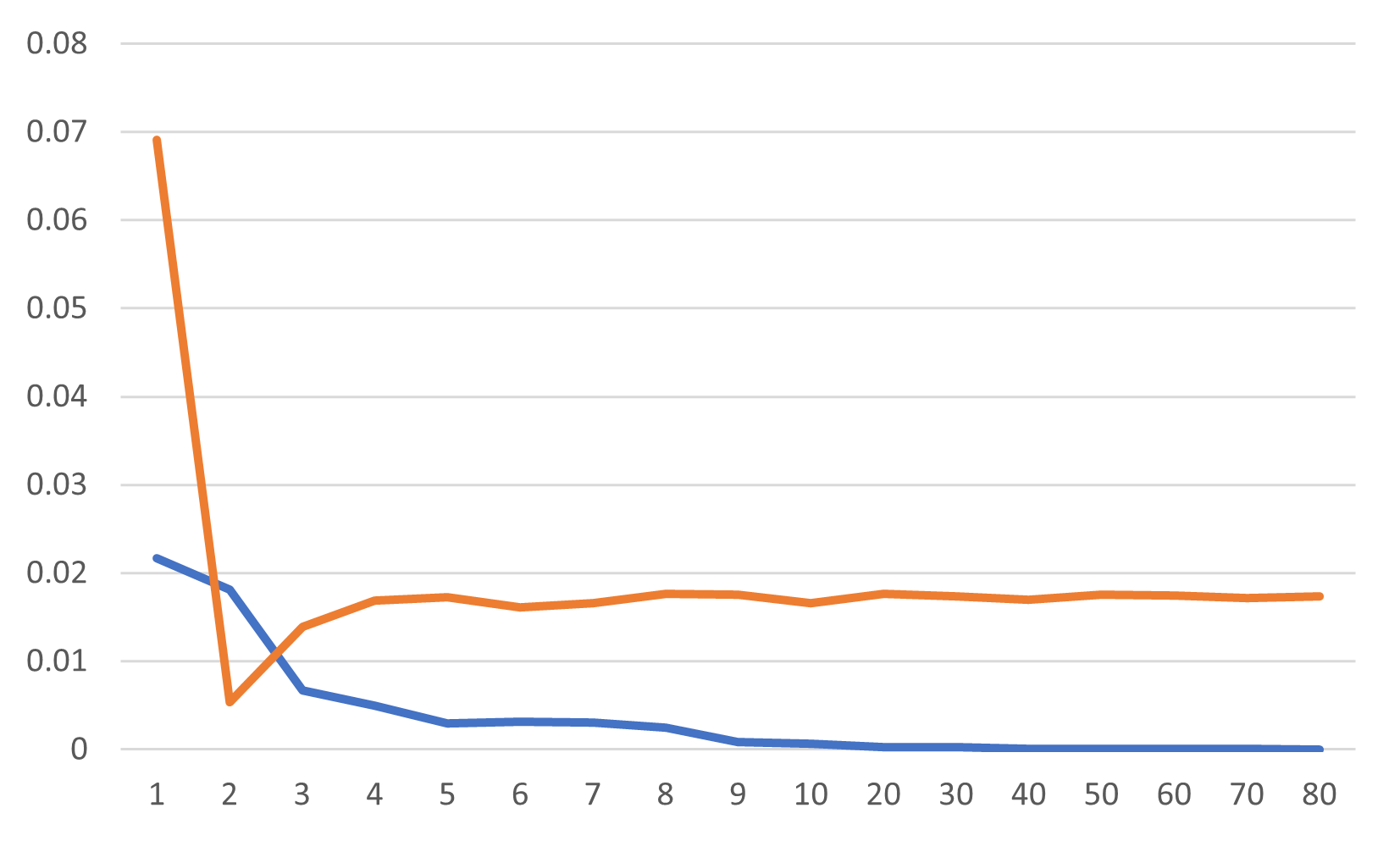}
\caption{Fruit-360: Comparing wrong verified cases of various number of classifiers for MASKS and Voting.}
\label{fig:Fruit-360-Err-vs-voting}
\end{figure}

\begin{rem}(Comparing with a voting system)
\label{sec:ComparingVS}
\end{rem}
Although the primary purpose of MASKS is to verify property for a multi-classifier, we developed a ``major voting'' system to explain how our method can be significant in error reduction \cite{auda1995voting}. Here, classifiers are the same as those applied in examples \ref{exam2}, \ref{exam3}, and \ref{exam4}. In fig \ref{fig:Fashion-MNIST-Err-vs-voting}, \ref{fig:MNIST-Err-vs-voting}, and \ref{fig:Fruit-360-Err-vs-voting} a voting system with the same classifiers is compared with the developed MASKS algorithm. In the Major Voting system, unverified cases have more than one class with major votes; others are counted as verified. As it can be seen, MASKS's error reduction is more significant. It can show that interpreting the knowledge of multiple classifiers in a classification problem with distributed knowledge seems to have better results than major voting in these cases (i.e., in table \ref{tab:results}, for the Fashion-MNIST dataset, for 1000 classifiers, the number of errors in the MASKS is 4-times less than the voting system).

\begin{table}[]
    \centering
\begin{tabular}{ |c||c|c|c|c|c|  }
 \hline
 Dataset   & number of classifiers  & \multicolumn{2}{|c|}{verified} & unverified  & Method \\
   &   & correct  &  Wrong &  &\\
 \hline
 Fashion-MNIST   & 1    &9821&   89&  90 & MASKS\\
 Fashion-MNIST   & 1000    &9588&   17&  395 & MASKS\\
 Fashion-MNIST   & 1    &9892&   108&  0 & Major Voting\\
 Fashion-MNIST   & 1000    &9936&   64&  0 & Major Voting\\
 \hline
 MNIST&   1  & 9556   &71&  373 & MASKS\\
 MNIST&   608  & 9519   &3&  478 & MASKS\\
 MNIST&   1  & 9895   &105&  0 & Major Voting\\
 MNIST&   608  & 9949   &51&  0 & Major Voting\\
 \hline
 Fruit-360 & 1 & 10730&  233&  11725 & MASKS\\
 Fruit-360 & 74 & 15850&  0&  6838 & MASKS\\
 Fruit-360 & 1 & 21221&  1467&  0 & Major Voting\\
 Fruit-360 & 74 & 22288&  386&  14 & Major Voting\\
 \hline
\end{tabular}
    \caption{Comparing MASKS with Major Voting method for single and multi-classifiers scenarios }
    \label{tab:results}
\end{table}

\section{Conclusion and Future Works}
\label{sec:Conclusion}
Classification is one of the primary tasks of Artificial Intelligence (AI).
Nowadays, many classifiers are 
applied in critical applications, in which errors would cause serious impacts.    
This study aims to develop a multi-agent verification method to reduce errors by integrating multiple classifiers. Here, it is shown that a multi-agent scenario could perform better in order to reduce errors. To do this, primarily, a property has been defined to present the knowledge of the classifiers. 
Next, a multi-agent system was designed in order to investigate these multiple classifiers. Then, a dynamic epistemic logic-based method was developed for reasoning about the aggregation of distributed knowledge. This knowledge was acquired through separate classifiers and external information sources. 
Finally, it was shown that aggregated knowledge might lead to refining output results. As a result, the model could verify the model for a specific input, if the knowledge of the entire system satisfies its correctness. In other words, the system could distinguish robust answers.  To conclude, a multi-agent system for the knowledge sharing (MASKS) algorithm has been proposed for the aforementioned model. This proposed method was applied to the Fashion-MNIST, MNIST, and Fruit-360 datasets. As a result, the error rate of the entire system dropped significantly. 

Looking into the future, we aim to develop an approach that can model time-series classifiers (i.e., for \textit{recurrent neural networks} or \textit{reinforcement learning}) for real-time verifying approaches. Moreover, we will develop a tool to verify any multi-agent system's inputs and check whether an input point can be verified in the system. This tool should be able to manipulate knowledge sharing in trusted or untrusted networks. To enhance the performance of this tool, fuzzy logic could be applied to avoid state space explosion for classifiers, especially where more than two output classes exist. 


\section{Bibliography styles}

\bibliography{mybibfile}

\appendix
\section{MASKS's Tool}
\label{sec:appendixMTool}
MASKS is a tool in order to verify multi-classifier with a predefined property. Using MASKS, the satisfaction of the property for all classifiers would be checked. In other words, if the property $\varphi$ is satisfied using knowledge of all agents, it would be the verified formula (property). Here, we use the operator $D_A \varphi$ from Public Announcement Logic to collect distributed knowledge of classifiers. 

Here, for more convenience, we developed python codes in three steps to run the tool. Self-created ones could replace the first two steps. These two steps are developed to provide inputs for the MASKS tool. 

The first one is ``\texttt{0-create\_model.py}'', in which classifiers would be created. After defining the target dataset, the input data would be collected using the ``load\_input\_data'' function. The output of this function should contain a set of train images, and it should be stored in ``train\_data'' variable. In this python file, the train data would be applied to train classifiers (i.e.,  ANNs). The architecture of the classifiers could be defined in ``define\_model'' function in ``\texttt{build\_model.py}''. The number of output classes (``no\_classes'') and the input shape (``input\_shape'') should be determined. The output would be multiple classifiers. These classifiers would be stored into the ``Models'' folder. The number of classifiers could be determined by ``model\_no'' variable.

In order to run ``\texttt{0-create\_model.py}'', following inputs are required:
\begin{itemize}
    \item The number of output classes:
    would be stored in ``no\_classes'' folder,
    \item The dimension of input images:
    would be stored in ``input\_shape'' folder,
    \item The number of agents to be created:
    would be stored in ``model\_no'' folder,
    \item train and validation dataset: would be stored in ``train\_data'' and ``validation\_data'' folders (validation is optional),
    \item The classifier architecture:
    could be defined in ``define\_model'' folder in ``\texttt{build\_model.py}'' file,
    \item If you using ``\texttt{load\_input\_data.py}'' for loading image files, the ``train\_df\_path'' should be set to be the train file path.
\end{itemize}

Next, using the stored model in the ``Models'' folder, ``\texttt{1-Eval\_model.py}'' could be executed to evaluate test inputs and their neighbourhoods and manipulations. The set of neighbourhoods and manipulations would be defined in \textit{python class} ``NeighMan'' (here is a set of noise on the input image). The output of ``\texttt{1-Eval\_model.py}'' would be the results of inputs, neighbourhoods, and manipulations for each classifier in a \textit{numpy} file in ``Results'' folder.

In order to run ``\texttt{1-Eval\_model.py}'', following inputs are required:
\begin{itemize}
    \item Image dimension:  would be stored in ``img\_width'', ``img\_height'', and ``img\_num\_channels'' folders,
    \item The input images: would be stored in ``input\_test'' folder,
    \item Outputs of test inputs (and their neighbourhoods and manipulations) for each classifier: would be stored \textit{numpy} array in files ``Results'' folder,
    \item The number of neighbourhoods and manipulations:
    would be stored in ``no\_classes'' folder,
    \item Ordered correct labels of inputs: would be stored in ``target\_test'' folder,
    \item The ``project\_path'', ``data\_dir'',  should be set to be the project path and test file path.
\end{itemize}

Then, using the output results in the ``Results'' folder, ``\texttt{2-MASKS.py}'' could be executed. Here, the correct output labels would be provided with the function ``load\_labels''. After execution of this python code, the following results for one to the number of classifiers would be provided:
\begin{enumerate}
    \item ``agent\_counter'': number of agents,
    \item ``correct\_answer'': correct verified answers,
    \item ``wrong\_answer'': wrong verified answer,
    \item ``conflict\_answer'': unverified answers because of conflicting,
    \item ``correct\_assist'': unverified answers because more than one output class is provided, but the correct answer is in the provided output classes,
    \item ``wrong\_assist'':  unverified answers because more than one output class is provided, but the correct answer is not in the provided output classes.
\end{enumerate}
For further investigation, the result of these multi-agent systems, every input would be stored in the ``Agents'' folder.

In order to run ``\texttt{2-MASKS.py}'', following inputs are required:
\begin{itemize}
    \item Outputs of test inputs (and their neighbourhoods and manipulations) for each classifier: would be stored \textit{numpy} array in files ``Results'' folder,
    \item The number of output classes:
    would be stored in ``no\_classes'' folder,
    \item The number of neighbourhoods and manipulations:
    would be stored in ``nei\_man\_no'' folder,
    \item Ordered correct labels of inputs: would be stored in ``target\_test'' folder.
\end{itemize}

The tool and Modified Fashion MNIST, MNIST, and Fruit-360 could be found in \texttt{https://github.com/iuwa/MASKS} (examples are \texttt{FashionMNIST-MASKS.zip}, \texttt{MNIST-MASKS.zip}, and \texttt{Fruit-360-MASKS.zip}).
\end{document}